\documentclass[lettersize,journal]{IEEEtran}
\usepackage{amsmath,amsfonts}
\usepackage{algorithmic}
\usepackage{algorithm}
\usepackage{array}
\usepackage[caption=false,font=normalsize,labelfont=sf,textfont=sf]{subfig}
\usepackage{textcomp}
\usepackage{stfloats}
\usepackage{url}
\usepackage{verbatim}
\usepackage{graphicx}
\usepackage{cite}
\usepackage{multirow}
\usepackage{color}
\usepackage{colortbl}
\usepackage{booktabs}
\definecolor{RowColor}{rgb}{0.95, 0.95, 1}
\begin{document}

\bibliographystyle{IEEEtran}

\title{Factorization Vision Transformer: Modeling Long Range Dependency with Local Window Cost}

\author{Haolin Qin$^{*}$, Daquan Zhou$^{*}$, Tingfa Xu$^{\dagger}$, Ziyang Bian, Jianan Li$^{\dagger}$
\thanks{Haolin Qin, Tingfa Xu, Jianan Li and Ziyang Bian are with Beijing Institute of Technology, 100081 Beijing, China. E-mail:\{3120225333, ciom\_xtf1, lijianan, 3220185049\}@bit.edu.cn.}
\thanks{Tingfa Xu is also with Beijing Institute of Technology Chongqing Innovation Center, 401135 Chongqing, China.}
\thanks{Daquan Zhou is with ByteDance TikTok, Singapore.}
\thanks{Ziyang Bian is also with North China Research Institute of Electro-Optics, 100015 Beijing, China.}
\thanks{$^{*}$ Equal contribution. $^{\dagger}$ Correspondence to: Tingfa Xu and Jianan Li.}}

\markboth{Journal of \LaTeX\ Class Files,~Vol.~14, No.~8, August~2021}%
{Qin \MakeLowercase{\textit{et al.}}: Factorization Vision Transformer: Modeling Long Range Dependency with Local Window Cost}


\maketitle

\begin{abstract}
Transformers have astounding representational power but typically consume considerable computation which is quadratic with image resolution. The prevailing Swin transformer reduces computational costs through a local window strategy. However, this strategy inevitably causes two drawbacks: (1) the local window-based self-attention hinders global dependency modeling capability; (2) recent studies point out that local windows impair robustness. To overcome these challenges, we pursue a preferable trade-off between computational cost and performance. Accordingly, we propose a novel factorization self-attention mechanism (FaSA) that enjoys both the advantages of local window cost and long-range dependency modeling capability. By factorizing the conventional attention matrix into sparse sub-attention matrices, FaSA captures long-range dependencies while aggregating mixed-grained information at a computational cost equivalent to the local window-based self-attention. Leveraging FaSA, we present the factorization vision transformer (FaViT) with a hierarchical structure. FaViT achieves high performance and robustness, with linear computational complexity concerning input image spatial resolution. Extensive experiments have shown FaViT's advanced performance in classification and downstream tasks. Furthermore, it also exhibits strong model robustness to corrupted and biased data and hence demonstrates benefits in favor of practical applications. In comparison to the baseline model Swin-T, our FaViT-B2 significantly improves classification accuracy by $1\%$ and robustness by $7\%$, while reducing model parameters by $14\%$. Our code will soon be publicly available at https://github.com/q2479036243/FaViT.
\end{abstract}


\section{Introduction}
\IEEEPARstart{S}{ince} the great success of Alexnet \cite{krizhevsky2012imagenet}, revolutionary improvements have been achieved through scaling the model size to a larger scale with several advanced training recipes \cite{he2019bag}. With the aid of AutoML \cite{zoph2016neural}, convolutional neural networks (CNNs) have achieved state-of-the-art performance across various computer vision tasks \cite{zhang2023fast, tan2019efficientnet}. On the other hand, recently popular transformers have shown superior performance over previously dominant CNNs \cite{liu2023survey, kaselimi2022vision}. The fundamental difference between transformers and CNNs resides in their aptitude for modeling long-range dependency. Conventional vision transformers divide input images into sequences of patches, processed concurrently, thereby yielding a quadratic increase in computational cost with respect to input spatial resolution. As a result, transformers consume significantly higher computational costs compared to CNNs, limiting their feasibility in resource-constrained devices.

\begin{figure}[!t]
\centering
\includegraphics[width=0.8\linewidth]{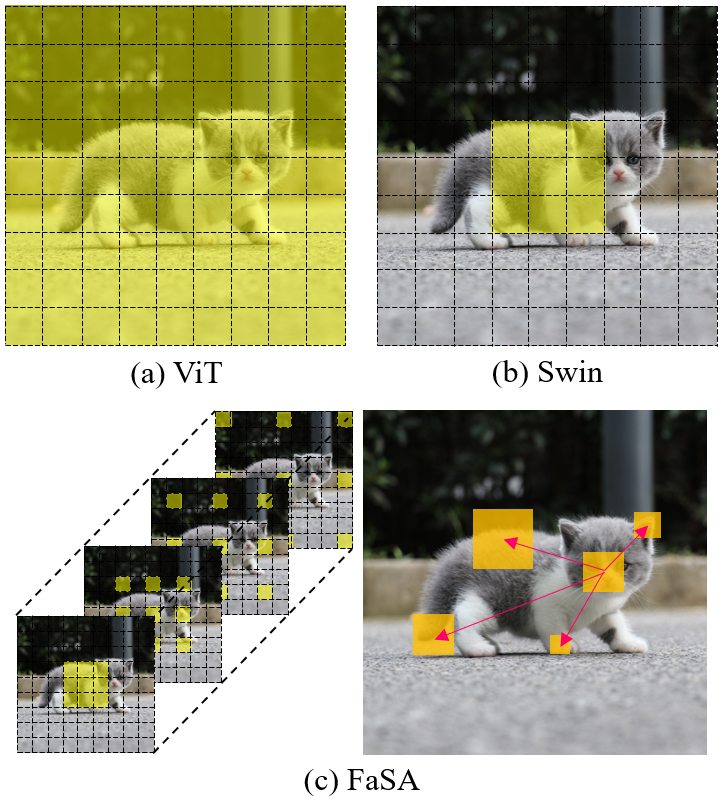}
\caption{Comparison of the attention span visualization results of ViT, Swin, and our FaViT. (a) ViT enjoys a global attention span but is computationally intensive. (b) Swin transformer is efficient but inferior in modeling long-range dependency. (c) The proposed FaViT factorizes tokens and hence successfully models long-range dependency at local window cost.}
\label{fig:atten}
\end{figure}

To solve the above problem, several methods have been proposed to alleviate the computational burden. For example, the Swin transformer \cite{liu2021swin} adopts a local window strategy where tokens are divided into several windows, and the self-attention calculation is confined within these predefined windows. Consequently, the computational cost changes to be quadratic with the window size, which is intentionally set to be significantly smaller than the input image spatial resolution. Compared with conventional vision transformer (ViT) \cite{dosovitskiy2020image}, the Swin transformer reduces costs significantly. Nonetheless, this local window strategy inevitably impairs the long-range dependency modeling capability. As illustrated in Figure~\ref{fig:atten} (b), the Swin transformer exclusively captures relationships within a local area, leading to the loss of long-range dependency. Additionally, the damage of this strategy to model robustness has been demonstrated by recent investigations \cite{zhou2022understanding}.

\begin{figure}[!t]
\centering
\includegraphics[width=3in]{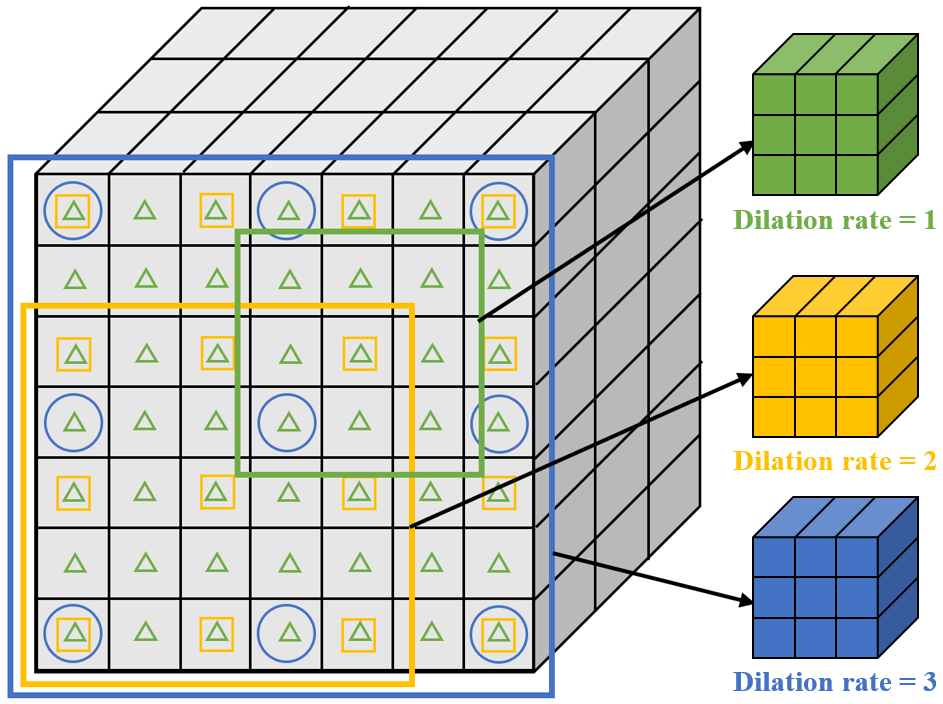}
\caption{The factorization with multiple dilation rates. We factorize the conventional attention matrix into several sub-attention matrices and obtain long-range and mixed-grained information with computational complexity linearly with respect to the image resolution.}
\label{fig:dila}
\end{figure}

The trade-off between the computational cost and long-range dependency modeling capability thus becomes a fundamental challenge yet to be explored. In this paper, we propose a novel self-attention mechanism termed factorization self-attention (FaSA). The core operation of FaSA lies in the factorization process, illustrated in Figure \ref{fig:dila}. We factorize the conventional attention matrix into several sparse sub-attention matrices such that the super-position of sub-attention matrices effectively approximates the original full attention matrix. 

In practical terms, given an input image, we take each individual point as a query. For gathering keys, we evenly divide the image into a series of non-overlapping local windows. Subsequently, we uniformly sample a fixed number of points from each window through dilated sampling and fuse the features of the sampled points at the same position in different windows. Since the number of keys is strictly limited and each key incorporates information spanning various windows across the entire image, attending to such a set of keys enables modeling long-range dependency at local window cost. Considering that each obtained key amalgamates multi-point information, potentially resulting in a deficiency of fine-grained details, we further introduce the mixed-grained multi-head attention. Concretely, we incrementally increase the local window size for different heads and dynamically adjust the dilation rate of point sampling to keep the number of keys across all heads the same. As a result, the obtained keys incorporate features from fewer positions and have more fine-grained details without adding additional computing costs. By aggregating the attended features from multiple heads, long-range and mixed-grained information can be obtained simultaneously.

\begin{figure}[!t]
\centering
\includegraphics[width=3.5in]{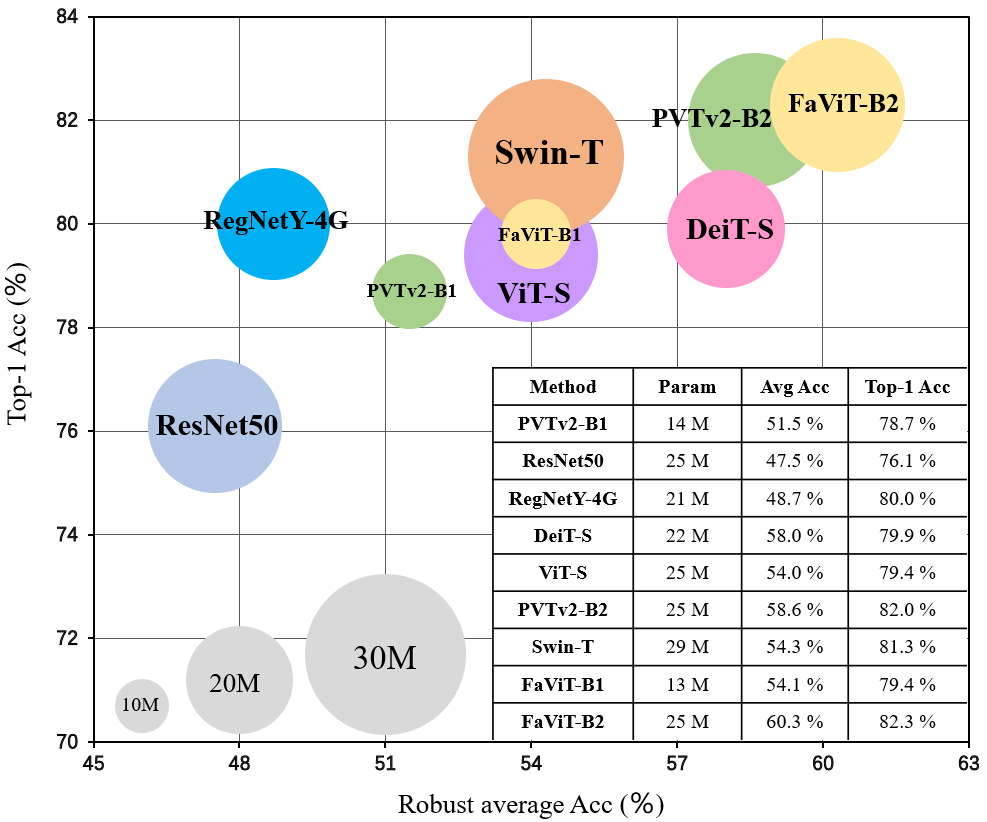}
\caption{Comparison of the accuracy with respect to robustness trade-off. The proposed FaViT achieves the best performance in both classification accuracy and model robustness, with fewer parameters which are indicated by the circle sizes.}
\label{fig:mce}
\end{figure}

Based on the proposed FaSA, we present variants of the transformer with different capacities termed factorization vision transformer (FaViT). As shown in Figure \ref{fig:atten} (c), our FaViT enjoys two essential advantages, stemming from the innovation of FaSA, which are previously unattainable in existing transformers. Firstly, each local window generates an identical number of tokens, ensuring a fixed computational cost. Consequently, the capture of long-range dependency transpires without incurring supplementary overhead. Secondly, the incorporation of sub-attention matrices within FaViT fosters the aggregation of information at varied granularities. 

Extensive experiments corroborate the exceptional performance and superior robustness of the proposed FaViT. As depicted in Figure \ref{fig:mce}, FaViT consistently outperforms competing models of similar sizes in terms of classification accuracy and robustness. Notably, in comparison to the baseline model Swin transformer, FaViT-B2 outperforms Swin-T in all aspects. The robustness is significantly improved by $7\%$, and the classification accuracy is improved by $1\%$, while the parameters drop by a considerable $14\%$. Furthermore, our FaViT also exhibits impressive performance on multiple downstream tasks such as object detection and semantic segmentation. 

To sum up, this work makes the following contributions: 
\begin{itemize}
\item{We propose a novel factorization self-attention mechanism (FaSA), which is capable of modeling long-range dependency while aggregating mixed-grained information at local window cost.}
\item{Based on FaSA, we present an efficient factorization vision transformer (FaViT), which finds the trade-off between cost and performance and exhibits state-of-the-art robustness.}
\end{itemize}

\section{Related work}
\subsection{Vision Transformers} 
Transformer is originally developed for NLP tasks \cite{brown2020language, devlin2018bert}, but has now achieved remarkable success across multiple domains \cite{wang2023learning, chen2022hider, zhao2022fractional, Cui2022MixFormerET}. In the computer vision tasks, ViT \cite{dosovitskiy2020image} is the pioneering attempt to adapt transformers for image processing. It splits the input image into a sequence of tokens for encoding and constructs a convolution-free network structure through self-attention. DeiT \cite{touvron2021training} introduces a series of training strategies to make ViT work on the smaller dataset ImageNet-1K, replacing the large-scale JFT-300M previously used. Building upon ViT's foundation, subsequent endeavors \cite{Jiang2021AllTM, Ding2022DaViTDA, Tu2022MaxViTMV} refine its architecture to achieve enhanced performance. For instance, Yu et al. \cite{yu2022metaformer} abstracted the general MetaFormer architecture and explored the sources of competitiveness of transformers and their variants. Jiao et al. \cite{jiao2023dilateformer} used sparse receptive fields to reduce global redundancy and improve transformer feature expression capabilities. In essence, the transformative impact of transformers has been extended to the domain of computer vision and gradually replaced the dominance of CNN.

\subsection{Efficient Variants} 
A recent surge of methods has emerged to mitigate the computational demands of vision transformers. These methods can be categorized into two strategies according to the relationship between computational complexity and the input image spatial resolution: (1) pyramid-based self-attention; and (2) local window-based self-attention. PVTv1 \cite{wang2021pyramid} serves as a representative example of the first strategy.
It introduces spatial reduction and tokens fusion to reduce the cost of multi-head attention. The subsequently proposed PVTv2 \cite{wang2022pvt} improves it by introducing overlapping patch embedding, depth-wise convolution \cite{chollet2017xception}, and linear Spatial Reduction Attention. A shunted self-attention \cite{ren2021shunted} comes up to unify multi-scale feature extractions via multi-scale token aggregation. The key point of this strategy is to reduce cost through spatial compression. Therefore, the computational complexity of the aforementioned pyramid-based self-attention models remains essentially quadratic with the image resolution. The high computational cost problem still exists when processing high-resolution images. Meanwhile, the information of small targets and delicate textures will be overwhelmed, destroying the fine-grained features.

\subsection{Local Self-attention} 
The local window-based self-attention divides the image into distinct subsets and confines self-attention computations within each subset. Therefore, the computational complexity of models using local windows is theoretically linear with the input image spatial resolution. Among these models, The most Swin transformer \cite{liu2021swin}  stands out as the most prominent. It introduces a hierarchical sliding window structure and utilizes the Shift operation to exchange information across subsets. MOA \cite{patel2022aggregating} exploits the neighborhood and global information among all non-local windows. SimViT \cite{li2021simvit} integrates spatial structure and cross-window connections of sliding windows into the visual transformer. While the local window strategy significantly reduces the computational cost, its drawbacks cannot be ignored. The absence of long-range dependency restricts representational capacity, while excessive local windows will compromise model robustness. Previous works are unable to restore long-range dependency while maintaining the cost. These issues drive us to propose FaViT for modeling long-range dependency with the local window-based self-attention computational cost.

\subsection{Model Robustness}
Compared with CNNs, transformers normally have stronger model robustness against various corruptions and biases, due to the utilization of the self-attention mechanism \cite{Bai2021AreTM, Xie2021SegFormerSA}. However, some transformer-based variants use strategies intended to enhance performance and reduce cost, which inadvertently damage model robustness \cite{zhou2022understanding}. Hence, we have integrated model robustness as a fundamental criterion within our performance evaluation framework. Robustness evaluation no longer relies solely on clean image datasets such as ImageNet-1K \cite{Paul2022VisionTA}. Instead, the assessment is based on ImageNetC which denotes the corrupted images derived from the original dataset in a manner akin to the approach proposed in \cite{Hendrycks2019BenchmarkingNN}. In addition, the model's performance against label noise and class imbalance also serves as an indicator of its robustness. This perspective is endorsed by Li et al. \cite{Li2020DivideMixLW}, who employed images with real-world noisy labels, and Tian et al. \cite{Tian2021VLLTRLC} used a long-tailed distribution dataset to analyze model robustness through classification accuracy. Drawing inspiration from these contributions, our comprehensive evaluation framework gauges the model's robustness across image corruption, label noise, and class imbalance scenarios.

\subsection{Matrix Factorization}
Recently, the construction of sparse matrices to reduce network redundancy and enhance feature extraction efficiency has emerged as a promising approach across various tasks \cite{che2021bicriteria, liu2021non, ren2020parallelized}. This approach has inspired the formulation of strategies aimed at reducing transformer computational costs. One notable instance is VSA \cite{zhang2022vsa}, which introduces a window regression module to predict the attention region where key and value tokens are sampled. This category of methods endeavors to find sparse tokens that can effectively substitute all tokens. Besides that, dilated sampling is also an effective strategy for achieving sparse representation \cite{zhang2023ctfnet, fu2020deep, liu2022self}. For example, Jiao et al. \cite{jiao2023dilateformer} devised DilateFormer, which introduces an additional dilation rate in the sliding window of the Swin transformer. These methods leverage dilated convolutions to extract sparse feature maps, thereby expanding the receptive field while reducing computational costs. However, dilated sampling will lose information continuity, which is neglected by the above methods, resulting in the loss of fine-grained features. To this end, this paper proposes FaSA to achieve attention matrix factorization and obtain long-range and mixed-grained information simultaneously.

\section{Method}

In this section, we present a comprehensive overview of the proposed factorization vision transformer (FaViT). In Section \ref{3.1}, we review the fundamentals of the transformer. We provide a main symbol comparison table facilitating a clearer understanding of the main differences between FaViT and existing models. Section \ref{3.2} is dedicated to the elucidation of our core innovation, the factorization self-attention (FaSA). This novel mechanism empowers FaViT to model long-range dependency at the local window cost. In Section \ref{3.3}, we describe the overall framework and model variants of the FaViT. Section \ref{3.4} is dedicated to a meticulous analysis of FaViT's computational complexity.

\begin{figure*}[!t]
\centering
\includegraphics[width=7in]{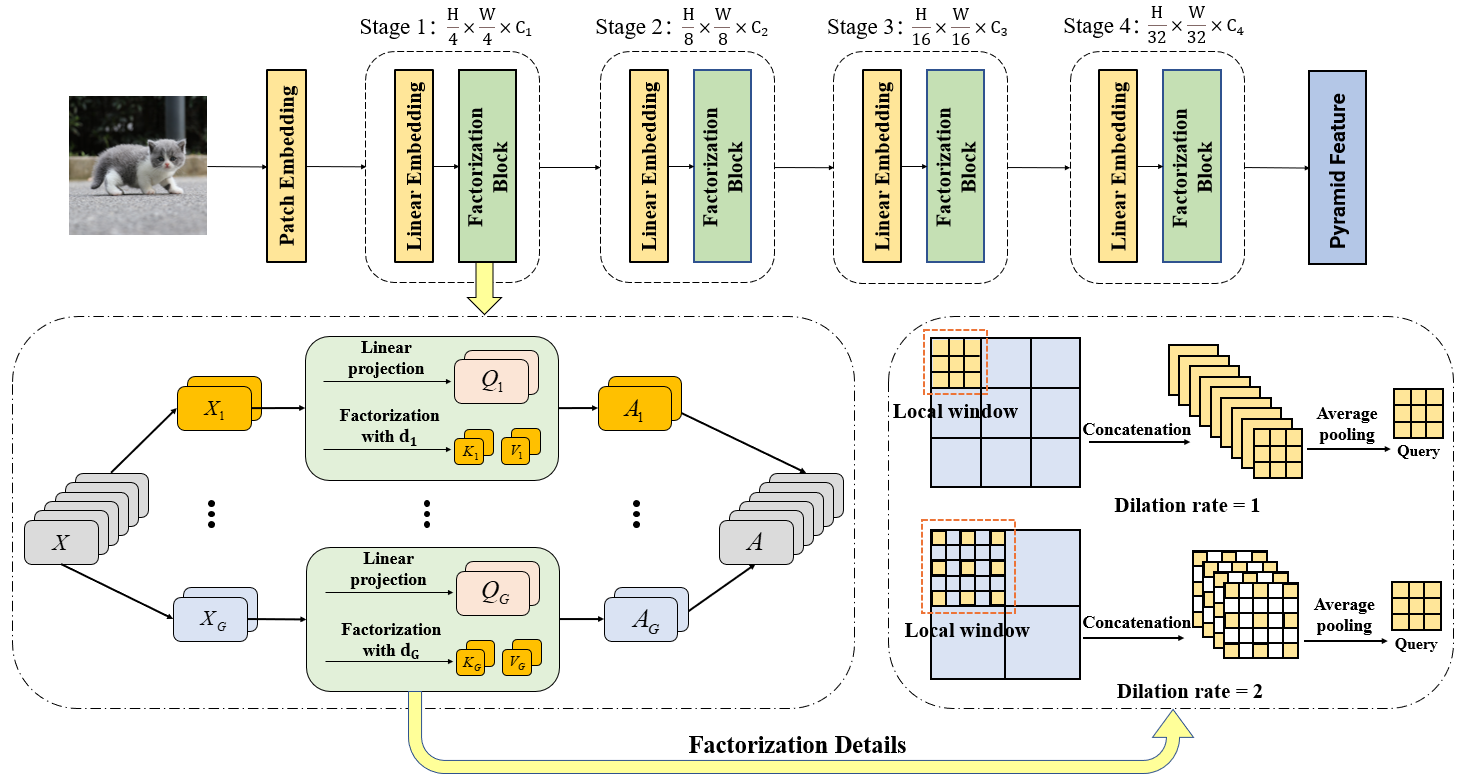}
\caption{Overall architecture of the proposed FaViT. We divide the attention heads into multiple groups and assign increasing dilation rates. Each group is evenly partitioned into non-overlapping local windows and gathers the same number of keys via dilated sampling to obtain the long-range and mixed-grained information simultaneously.}
\label{fig:all}
\end{figure*}

\subsection{Preliminaries}
\label{3.1}
Given the input image ${\bf{I}} \in \mathbb{R}^{\rm{H \times W \times 3}}$, where $\rm{H,W}$ represent its height and width respectively. The conventional self-attention mechanism (SA) employed in ViT first encodes it into a feature map ${\bf{X}} \in \mathbb{R}^{\rm{N \times C}}$ by patch embedding, where $\rm{N = H \times W}$ represents the input image spatial resolution and $\rm{C}$ represents its channel dimensions. Subsequently, SA employs linear embeddings, parameterized by weight matrices ${\bf{W}}^{\rm{K}}, {\bf{W}}^{\rm{Q}}, {\bf{W}}^{\rm{V}} \in \mathbb{R}^{\rm{C \times C}}$, to embed all the points into key ${\bf{K}} =  {\bf{W}}^{\rm{K}} {\bf{X} \in \mathbb{R}^{\rm{N \times C}}}$, query ${\bf{Q}} =  {\bf{W}}^{\rm{Q}} {\bf{X} \in \mathbb{R}^{\rm{N \times C}}}$, and value ${\bf{V}} =  {\bf{W}}^{\rm{V}} {\bf{X} \in \mathbb{R}^{\rm{N \times C}}}$, respectively. After that, the attention feature map is generated as indicated by the following equation:
\begin{equation}
\label{eqn:sa}
\text{SA}({\bf{X}}) = \text{Softmax}(\frac{{\bf{QK}}^{\top}}{\sqrt{h}}) {\bf{V}},
\end{equation}
where $\sqrt{h}$ is a scaling factor. As a result, the computational complexity of ViT is given by:
\begin{equation}
\label{eqn:sa_cost}
\Omega({\rm{ViT}}) = 4 \rm{N C^2 + 2 N^2 C},
\end{equation}
and this complexity grows quadratically in tandem with the increase in spatial resolution of the input image. Finally, transformers transform the attended features via adopting a feedforward layer, generating the ultimate feature maps that are tailored for specific visual tasks. Notably, most feedforward layers are based on Multilayer Perceptron (MLP), usually consisting of two linear layers and a GELU layer.

The Swin transformer adopts local window-based self-attention (WSA) to reduce the computational cost. WSA divides the feature map ${\bf{X}}$ into non-overlapping windows and performs self-attention calculations within each window in isolation. Suppose each window contains $\rm{M \times M}$ tokens, the computational complexity of the Swin transformer can be expressed as:
\begin{equation}
\label{eqn:wsa_cost}
\Omega({\rm{Swin}}) = 4 \rm{N C^2} + 2 M^2 \rm{NC}.
\end{equation}
Notably, this complexity exhibits linear growth with respect to the input resolution for a fixed $\rm{M}$.

A comparison between Equation \ref{eqn:sa_cost} and Equation \ref{eqn:wsa_cost} highlights that SA enjoys a global attention span but becomes computationally intensive for high input resolutions. In contrast, WSA achieves higher efficiency while sacrificing the ability to model long-range dependency, consequently affecting performance and robustness. To address these limitations, the Swin transformer introduces the Shift operation to facilitate information interaction among adjacent windows. However, the Shift operation is employed exclusively between WSA blocks, and expanding the modeling range necessitates the connection of multiple blocks in series. The above constraints prompt us to explore a novel self-attention mechanism that can effectively model long-range dependency while adhering to the computational advantages of local window strategy. It's worth noting that the main symbols used throughout this explanation are summarized in Table \ref{Symbol}, providing a reference for the core concepts introduced.

\begin{table}[!t]
\caption{Main Symbol Comparison Table.\label{Symbol}}
\centering
\renewcommand\arraystretch{1.2}
\setlength{\tabcolsep}{4pt}
\begin{tabular}{|ll|ll|}
\hline
Typeset & Symbol Description & Typeset & Symbol Description \\
\hline
I   &   Input image     &   X   &   Embedding feature\\
K   &   Key tokens      &   Q   &   Query tokens\\
V   &   Value tokens    &   $\rm{X}^{'}$  &   Attention feature\\
$\rm{F}$  &   Output feature    &   $\rm{W^K}$  &   Key linear embedding\\
H   &   Image height    &   W   &   Image width\\
N   &   Image spatial resolution &  C   &   Image channel dimension\\
G   &   Group number    &   $\rm{M}^2$   &   Window number\\
h   &   Scaling factor  &   ${\rm{X}}_i$  &   Group feature\\
${\rm{K}}_{i}$  &   Group key   &   $\rm{X}^{'}_{i}$  &   Group attention feature\\
$\rm{W}^{\rm{K}}_{\emph{i}}$  &   Group key embedding   &   $\rm{P}_{\emph{i}}$  &   Group sampling points\\
$\rm{C}^{'}$ &   Group channel dimension    &   ${\rm{D}}_{i}$  &   Group dilation rate\\
${\rm{S}}_{i}$  &   Group window size   &   ${\rm{M}}_i$    &   Group window number\\
$h_i$   &   Group scaling factor    &   $\rm{F}_{\emph{i}}$  &   Stage output\\
$\rm{X}^{\emph{j}}_{\emph{i}}$  &   Window feature  &   $\rm{P}^{\emph{j}}_{\emph{i}}$  &   Window sampling points\\
$\Omega$  &   Computational complexity  &   O   &   Positive correlation\\
\hline
\end{tabular}
\end{table}

\subsection{Factorization Self-attention}
\label{3.2}
Figure \ref{fig:all} provides a visual representation of the proposed factorization self-attention mechanism (FaSA). The essence of FaSA lies in the factorization of the conventional attention matrix into multiple sparse sub-attention matrices. This factorization methodology fosters the integration of features across diverse granularities, culminating in an approximation of the feature representation capacity exhibited by the complete attention matrix. Importantly, this approximation is achieved while adhering to the linear relationship between computational complexity and the spatial resolution of the input image.

Specifically, we first uniformly divide the input feature map ${\bf{X}} \in \mathbb{R}^{\rm{N \times C}}$ into multiple groups. Each group independently undergoes a self-attention process aimed at capturing long-range features at a specific granularity level. Regarding the self-attention in each group, we take all points of the feature map as query tokens and gather key tokens in three steps:
\begin{enumerate}
\item{\textit{Localization}, evenly divide the entire feature map into a series of local windows.}
\item{\textit{Dilated sampling}, uniformly sample a fixed number of points in each local window.}
\item{\textit{Cross-window fusion}, fuse the features of the sampled points at the same position in different windows.}
\end{enumerate}
Hence the resulting fused key tokens encapsulate long-range information from multiple windows spanned across the entire feature map.

To counter the potential loss of fine-grained details in the generated key tokens due to matrix sparsity, we introduce mixed-grained multi-head attention. This involves a progressive increase in the local window size across distinct head groups, coupled with an adaptive escalation in the corresponding dilation rate for point sampling. This step ensures a consistent count of sampled points despite the window size changes, resulting in key tokens with mixed-grained information. Next, we delve into the principle and implementation of each step in FaSA.

\subsubsection{Head Grouping}

We initiate the process by uniformly partitioning the input feature map ${\bf{X}}$ into multiple groups along the channel dimension:
\begin{equation}
\label{eqn:grouping}
{\bf{X}} = \{ {\bf{X}}_i \in \mathbb{R}^{\rm{N \times C}^{'} }, | i=1,\cdots,{\rm{G}} \},
\end{equation}
where $\rm{G}$ is the number of head groups, ${\bf{X}}_{i}$ is the feature map assigned to the $i$-th group, and $\rm{C^{'}=C/G}$. We take the divided features as inputs for distinct attention head groups, each executing factorization self-attention independently. This isolation facilitates the capture of long-range information across various granularities.

\subsubsection{Gathering Queries} The gathering of query (${\bf{Q}}$), key (${\bf{K}}$) and value (${\bf{V}}$) is executed individually within each attention head group. For the $i$-th head group, we treat each point of ${\bf{X}}_i$ as a query and obtain query features as:
\begin{equation}
\label{eqn:queries}
{\bf{Q}}_i = {\bf{W}}_i^{\rm{Q}} {\bf{X}}_i \in \mathbb{R}^{\rm{N \times C}^{'} },
\end{equation}
where ${\bf{W}}_i^{\rm{Q}} \in \mathbb{R}^{\rm{C^{'} \times C}^{'}}$ symbolizes a learnable linear embedding achieved through a $1 \times 1$ convolution operation. The resulting query matrix ${\bf{Q}}_i$ corresponds to the query features within the $i$-th head group. 

\subsubsection{Gathering Keys} The acquisition of keys largely determines the attention span and computational cost of a self-attention mechanism. In order to strike a balance between modeling long-range dependency and retaining the local window cost, we gather keys in the following three steps.

\textbf{Step 1: Local Windowing.} 

We commence by uniformly partitioning ${\bf{X}}_i$ into multiple non-overlapping local windows using a sliding window strategy. For the sake of simplicity, we assume $\rm{W = H}$, signifying that each local window possesses equal length and width denoted as $\rm{S}_\emph{i}$. This can be mathematically expressed as:
\begin{equation}
\label{eqn:keys1}
{\bf{X}}_i = \{ {\bf{X}}_i^j \in \mathbb{R}^{ \rm{S}_\emph{i} \times \rm{S}_\emph{i} \times \rm{C}^{'} }, | j=1,\cdots,\rm{M}_\emph{i} \}.
\end{equation}
Here $\rm{M}_\emph{i} = \rm{H} / \rm{S}_\emph{i}$ stands for the number of local windows within the $i$-th attention head group. ${\bf{X}}_i^j$ corresponds to the feature map of a specific local window. Notably, we gradually enlarge the size of each local window across various head groups, resulting in decreasing $\rm{M}_\emph{i}$.

\textbf{Step 2: Dilated Sampling.}

Subsequently, we proceed to uniformly sample a fixed number of $\rm{M \times M}$ points within each local window. In practice, we set $\rm{M}$ to $7$ by default. For the $i$-th head group, the sampled point set ${\bf{P}}_i$ is represented as:
\begin{equation}
\label{eqn:keys2}
{\bf{P}}_i = \{ {\bf{P}}_i^j \in \mathbb{R}^{ \rm{M \times M \times C}^{'} }, | j=1,\cdots,\rm{M}_\emph{i} \},
\end{equation}
where ${\bf{P}}_i^j$ denotes the sampled points for the $j$-th local window in the $i$-th group. Since different head groups have distinct local window sizes, it is critical to keep the number of points sampled across local windows in different head groups the same. In practice, we employ a concise but efficient dilated sampling with increasing dilation rates across head groups to achieve this objective. The corresponding dilation rate for the $i$-th head group $\rm{D}_\emph{i}$ is computed as: 
\begin{equation}
\label{eqn:keys3}
\rm{D}_\emph{i} = (\rm{S}_\emph{i} -1) / (\rm{M} -1).
\end{equation}
Consequently, the sampled points are uniformly distributed within each local window. As the head group index $i$ increases, the interval between sampled points widens, leading to the coarser granularity of information.

\begin{table*}[!t]
\caption{Architecture variants of FaViT. $\rm{P}_\emph{i}$, $\rm{C}_\emph{i}$, $\rm{H}_\emph{i}$, $\rm{E}_\emph{i}$, $\rm{B}_\emph{i}$, and $\rm{D}_\emph{i}$ indicate patch size, feature dimension, number of heads, expansion ratio of MLP, number of blocks, and dilation rate set respectively. \label{architecture}}
\centering
\renewcommand\arraystretch{1.5}
\setlength{\tabcolsep}{4pt}
\begin{tabular}{|c|c|c|cccc|}
\hline
 & Output Size & Layer Name & \multicolumn{1}{c|}{FaViT-B0} & \multicolumn{1}{c|}{FaViT-B1} & \multicolumn{1}{c|}{FaViT-B2} & \multicolumn{1}{c|}{FaViT-B3} \\
\hline
\multirow{3}{*}{Stage 1} & \multirow{3}{*}{$\rm{\frac{H}{4} \times \frac{W}{4}}$} & \multicolumn{1}{c|}{Linear Embedding} & \multicolumn{1}{c|}{$\rm{P_{1}=4; C_{1}=32}$} & \multicolumn{2}{c|}{$\rm{P_{1}=4; C_{1}=64}$} & \multicolumn{1}{c|}{$\rm{P_{1}=4; C_{1}=96}$} \\
\cline{3-7}
& & FaSA & \multicolumn{1}{c|}{$\begin{matrix} \rm{H_{1}=1} & \rm{E_{1}=8} \\ \rm{B_{1}=2} & \rm{D_{1}=[1,8]} \end{matrix}$} & \multicolumn{1}{c|}{$\begin{matrix} \rm{H_{1}=1} & \rm{E_{1}=8} \\ \rm{B_{1}=2} & \rm{D_{1}=[1,8]} \end{matrix}$} & \multicolumn{1}{c|}{$\begin{matrix} \rm{H_{1}=1} & \rm{E_{1}=8} \\ \rm{B_{1}=2} & \rm{D_{1}=[1,8]} \end{matrix}$} & \multicolumn{1}{c|}{$\begin{matrix} \rm{H_{1}=1} & \rm{E_{1}=8} \\ \rm{B_{1}=2} & \rm{D_{1}=[1,8]} \end{matrix}$} \\
\hline
\multirow{3}{*}{Stage 2} & \multirow{3}{*}{$\rm{\frac{H}{8} \times \frac{W}{8}}$} & \multicolumn{1}{c|}{Linear Embedding} & \multicolumn{1}{c|}{$\rm{P_{2}=2; C_{2}=64}$} & \multicolumn{2}{c|}{$\rm{P_{2}=2; C_{2}=128}$} & \multicolumn{1}{c|}{$\rm{P_{2}=2; C_{2}=192}$} \\
\cline{3-7}
& & FaSA & \multicolumn{1}{c|}{$\begin{matrix} \rm{H_{2}=2} & \rm{E_{2}=6} \\ \rm{B_{2}=2} & \rm{D_{2}=[1,4]} \end{matrix}$} & \multicolumn{1}{c|}{$\begin{matrix} \rm{H_{2}=2} & \rm{E_{2}=6} \\ \rm{B_{2}=2} & \rm{D_{2}=[1,4]} \end{matrix}$} & \multicolumn{1}{c|}{$\begin{matrix} \rm{H_{2}=2} & \rm{E_{2}=6} \\ \rm{B_{2}=3} & \rm{D_{2}=[1,4]} \end{matrix}$} & \multicolumn{1}{c|}{$\begin{matrix} \rm{H_{2}=2} & \rm{E_{2}=6} \\ \rm{B_{2}=3} & \rm{D_{2}=[1,4]} \end{matrix}$} \\
\hline
\multirow{3}{*}{Stage 3} & \multirow{3}{*}{$\rm{\frac{H}{16} \times \frac{W}{16}}$} & \multicolumn{1}{c|}{Linear Embedding} & \multicolumn{1}{c|}{$\rm{P_{3}=2; C_{3}=128}$} & \multicolumn{2}{c|}{$\rm{P_{3}=2; C_{3}=256}$} & \multicolumn{1}{c|}{$\rm{P_{3}=2; C_{3}=384}$} \\
\cline{3-7}
& & FaSA & \multicolumn{1}{c|}{$\begin{matrix} \rm{H_{3}=4} & \rm{E_{3}=4} \\ \rm{B_{3}=6} & \rm{D_{3}=[1,2]} \end{matrix}$} & \multicolumn{1}{c|}{$\begin{matrix} \rm{H_{3}=4} & \rm{E_{3}=4} \\ \rm{B_{3}=6} & \rm{D_{3}=[1,2]} \end{matrix}$} & \multicolumn{1}{c|}{$\begin{matrix} \rm{H_{3}=4} & \rm{E_{3}=4} \\ \rm{B_{3}=18} & \rm{D_{3}=[1,2]} \end{matrix}$} & \multicolumn{1}{c|}{$\begin{matrix} \rm{H_{3}=4} & \rm{E_{3}=4} \\ \rm{B_{3}=14} & \rm{D_{3}=[1,2]} \end{matrix}$}\\
\hline
\multirow{3}{*}{Stage 4} & \multirow{3}{*}{$\rm{\frac{H}{32} \times \frac{W}{32}}$} & \multicolumn{1}{c|}{Linear Embedding} & \multicolumn{1}{c|}{$\rm{P_{4}=2; C_{4}=256}$} & \multicolumn{2}{c|}{$\rm{P_{4}=2; C_{4}=512}$} & \multicolumn{1}{c|}{$\rm{P_{4}=2; C_{4}=768}$} \\
\cline{3-7}
& & FaSA & \multicolumn{1}{c|}{$\begin{matrix} \rm{H_{4}=8} & \rm{E_{4}=4} \\ \rm{B_{4}=2} & \rm{D_{4}=[1]} \end{matrix}$} & \multicolumn{1}{c|}{$\begin{matrix} \rm{H_{4}=8} & \rm{E_{4}=4} \\ \rm{B_{4}=2} & \rm{D_{4}=[1]} \end{matrix}$} & \multicolumn{1}{c|}{$\begin{matrix} \rm{H_{4}=8} & \rm{E_{4}=4} \\ \rm{B_{4}=3} & \rm{D_{4}=[1]} \end{matrix}$} & \multicolumn{1}{c|}{$\begin{matrix} \rm{H_{4}=8} & \rm{E_{4}=4} \\ \rm{B_{4}=3} & \rm{D_{4}=[1]} \end{matrix}$}\\
\hline
\end{tabular}
\end{table*}

\textbf{Step 3: Cross-window Fusion.}

Previous studies have demonstrated that confining self-attention exclusively within a local window detrimentally affects modeling capability and robustness, primarily due to the absence of cross-window information interaction \cite{liu2021swin}. To overcome this limitation while simultaneously reducing the number of key tokens, we introduce an innovative cross-window fusion strategy. Specifically, we first perform feature embedding for each sampled point. 
\begin{align}
{\bf{K}}_i^{'} &= {\bf{W}}_i^{\rm{K}} {\bf{P}}_i \in \mathbb{R}^{ \rm{M_\emph{i} \times M^2 \times  C}^{'} }, \\
{\bf{V}}_i^{'} &= {\bf{W}}_i^{\rm{V}} {\bf{P}}_i \in \mathbb{R}^{ \rm{M_\emph{i} \times M^2 \times  C}^{'} },
\end{align}
where ${\bf{W}}_i^{\rm{K}}, {\bf{W}}_i^{\rm{V}} \in \mathbb{R}^{\rm{C^{'} \times C}^{'}}$ denote learnable linear embeddings for the $i$-th group implemented by two separate $1 \times 1$ convolutions. ${\bf{K}}_i^{'}$ and ${\bf{V}}_i^{'}$ express the key and value in the $i$-th group. Subsequently, we fuse the features of sampled points situated at the same positions in different windows to obtain the final key and value features:
\begin{align}
{\bf{K}}_i &= \sigma({\bf{K}}_i^{'}) \in \mathbb{R}^{ \rm{M^2 \times C}^{'} }, \\
{\bf{V}}_i &= \sigma({\bf{V}}_i^{'}) \in \mathbb{R}^{ \rm{M^2 \times C}^{'} },
\end{align}
where $\sigma(\cdot)$ is a symmetric aggregation function. Hereby we implement it as a simple form, \textit{i.e.}, maximum pooling.

In the above process, the full attention matrix is factorized into several sparse sub-attention matrices. Each fused feature benefits from enriched long-range information, achieved by amalgamating the features of $\rm{M}_\emph{i}$ points distributed uniformly across the entire feature map. Moreover, as the head group index $i$ increases, the value of $\rm{M}_\emph{i}$ decreases, thereby enabling the fused feature to gather information from fewer positions. Consequently, it captures multiple granularity details. As a result, these sub-attention matrices can effectively approximate the feature representation performance of the full attention matrix.

\subsubsection{Mixed-grained Multi-head Attention} Given the gathered queries and keys for each head group, we perform self-attention individually:
\begin{equation}
\label{eqn:Mixed}
{\bf{X}}_i^{'} = \text{Softmax}(\frac{{\bf{Q}}_i {\bf{K}}_i^{\top}}{\sqrt{h_i}}) {\bf{V}}_i \in \mathbb{R}^{\rm{H \times W \times C}^{'}},
\end{equation}
where $\sqrt{h_i}$ denotes a scaling factor. Subsequently, we amalgamate the attended features from all head groups to obtain the final output. 
\begin{equation}
\label{eqn:Mixed1}
{\bf{X}}^{'} = \delta({\bf{X}}_i^{'}, | i=1,\cdots,\rm{G}) \in \mathbb{R}^{\rm{H \times W \times C}},
\end{equation}
where $\delta(\cdot)$ represents the concatenation operation along the feature channel dimension. Consequently, FaSA is capable of modeling long-range dependency while aggregating mixed-grained information.

\subsection{Overall Architecture}
\label{3.3}
Building upon the FaSA mechanism elucidated above, we propose the factorization vision transformer (FaViT) depicted in Figure \ref{fig:all}. Specifically, FaViT first feeds the input image $\bf{I}$ into a patch embedding layer to generate the feature map $\bf{X}$. Typically, patch embedding is executed using convolution to form non-overlapping patches. However, taking inspiration from recent works \cite{wang2022scaled,ren2021shunted}, this paper employs convolutional layers with a kernel size of $7 \times 7$ and a stride of $2$ to generate overlapping patches. Subsequently, a non-overlapping projection layer with a stride of 2 is employed to create the input feature map $\bf{X}$ with size of $\frac{\rm{H}}{4} \times \frac{\rm{W}}{4}$.

Following previous designs \cite{liu2021swin, wang2021pyramid}, we adopt a four-stage architecture to process $\bf{X}$ and generate hierarchical feature maps $\bf{F}$. The feature maps at each stage are denoted as ${\bf{F}} = \{ {\bf{F}}_i, | i=1,2,3,4 \}$, where ${\bf{F}}_i \in \mathbb{R}^{\rm{\frac{H}{2^{\emph{i}+1}} \times \frac{W}{2^{\emph{i}+1}} \times (C \times 2^{\emph{i}-1})}}$. These stages are interconnected by linear embedding layers, implemented using convolutional layers with a stride of 2. These layers reduce the size of the feature maps by half and double their dimension compared to the previous stage, which are then fed into the subsequent stage.

For each stage, we first stack multiple FaSA layers. The FaSA, as detailed above, employs local window and dilated sampling to obtain sparse sub-attention matrices. These sub-attention matrices facilitate the aggregation of information at multiple granularities through cross-window fusion. Consequently, they possess similar feature expression capabilities to the full attention matrix while simultaneously reducing computational costs. The calculated attention feature map $\bf{X}^{'}$ is then fed into a feedforward layer to generate the output feature map $\bf{F}_{\emph{i}}$ of the current stage. This feedforward layer is implemented as a Multilayer Perceptron (MLP), following the approach of previous work \cite{ren2021shunted}.

Based on the above network structure, we present four model variants of FaViT with distinct model sizes by adopting different parameter settings. 
More details of the model architecture can be found in table~\ref{architecture}.

\subsection{Complexity Analysis}
\label{3.4}
Our FaViT aims to achieve high performance and robustness with computational complexity linearly with respect to the input image spatial resolution. Since self-attention is computed within each local window, the computational complexity of FaViT is quadratic with respect to the window size and linear with respect to the image resolution. This complexity can be expressed as:
\begin{equation}
\label{eqn:cost}
\Omega({\rm{FaSA}}) = 4 \rm{N C^2 + 2 M^2 N C},
\end{equation}
where $\rm{M}$ is a preset fixed value, thus the computational complexity of FaViT is $O(\rm{N})$. In contrast, ViT has a complexity of $O(\rm{N}^2)$, and pyramid-based architectures have complexities of $O(\rm{N}^2/ \theta)$, where $\theta$ is an attenuation coefficient. The complexity of FaViT is theoretically significantly lower than the above methods. FaViT should naturally be classified into the Swin transformer category with higher performance and robustness. As a result, FaSA is capable of modeling long-range dependency at a computational cost equivalent to local window-based self-attention.

\section{Experiments}
The effectiveness and generalizability of our proposed FaViT have been evaluated across multiple computer visions. These evaluations encompass image classification, object detection, and semantic segmentation tasks. Furthermore, we assess FaViT's robustness through experimentation involving image corruptions, label noise, and class imbalance challenges. Ablation studies are provided to validate our design choices.

\subsection{Image Classification}
\label{IC}

\begin{table}[!t]
\caption{Classification results on ImageNet-1K. All models are trained from scratch using the same training strategies.\label{classification}}
\centering
\renewcommand\arraystretch{1.2}
\setlength{\tabcolsep}{4pt}
\begin{tabular}{c|l|ccc}
\toprule
Model Size & \multicolumn{1}{c|}{Method} & \#Param & FLOPs & Top-1 Acc (\%) \\
\midrule
\multirow{2}{*}{\begin{tabular}[c]{@{}l@{}}Tiny\\ Model\end{tabular}} & PVTv2-B0 \cite{wang2022pvt} & 4 M & 0.6 G & 70.5 \\ 
 & \cellcolor{RowColor}{FaViT-B0 (Ours)} & \cellcolor{RowColor}3 M & \cellcolor{RowColor}0.6 G & \cellcolor{RowColor}71.5 \\
\midrule
\multirow{5}{*}{\begin{tabular}[c]{@{}l@{}}Small\\ Model\end{tabular}} & ResNet18 \cite{he2016deep} & 12 M & 1.8 G & 69.8 \\
&PVTv1-T \cite{wang2021pyramid} & 13 M & 1.9 G & 75.1 \\
&PVTv2-B1 \cite{wang2022pvt} & 14 M & 2.1 G & 78.7 \\
&Mobile-Former \cite{chen2022mobile} & 14 M & 1.0 G & 79.3 \\
&\cellcolor{RowColor}{FaViT-B1 (Ours)} & \cellcolor{RowColor}13 M & \cellcolor{RowColor}2.4 G & \cellcolor{RowColor}79.4 \\
\midrule
\multirow{18}{*}{\begin{tabular}[c]{@{}l@{}}Medium\\ Model\end{tabular}} & DeiT-S \cite{touvron2021training} & 22 M & 4.6 G & 79.9 \\
&Distilled DeiT-S \cite{touvron2021training} & 22 M & 4.6 G & 81.2 \\
&NesT-T \cite{zhang2021aggregating} & 17 M & 5.8 G & 81.3 \\
&T2T-ViT-14 \cite{yuan2021tokens} & 22 M & 5.2 G & 81.5 \\
&ViL-S \cite{zhang2021multi} & 25 M & 5.1 G & 82.0 \\
&CrossViT-15 \cite{chen2021crossvit} & 27 M & 5.8 G & 81.5 \\
&TNT-S \cite{han2021transformer} & 24 M & 5.2 G & 81.5 \\
&DW-ViT-T \cite{ren2022beyond} & 30 M & 5.2 G & 82.0 \\
&DW-Conv-T \cite{han2021connection} & 24 M & 3.8 G & 81.3 \\
&GG-T \cite{yu2021glance} & 28 M & 4.5 G & 82.0 \\
&CoAtNet-0 \cite{dai2021coatnet} & 25 M & 4.2 G & 81.6 \\
&VSA \cite{zhang2022vsa} & 29 M & 4.6 G & 82.3 \\
&ConvNeXt-T \cite{Liu2022ACF} & 29 M & 4.5 G & 82.1 \\
&MViTv2-T \cite{li2022mvitv2} & 24 M & 4.7 G & 82.3 \\
&ViTAEv2-S \cite{zhang2023vitaev2} & 20 M & 5.4 G & 82.2 \\
&CrossFormer \cite{wang2023crossformer++} & 31 M & 4.9 G & 82.5 \\
&Swin-T \cite{liu2021swin} & 29 M & 4.5 G & 81.3 \\
& \cellcolor{RowColor}{FaViT-B2 (Ours)} & \cellcolor{RowColor}24 M & \cellcolor{RowColor}4.5 G & \cellcolor{RowColor}82.3 \\
\midrule
\multirow{18}{*}{\begin{tabular}[c]{@{}l@{}}Large\\ Model\end{tabular}} & LV-ViT-M \cite{Jiang2021AllTM} & 56 M & 16.0 G & 84.1 \\
&SimViT-M \cite{li2021simvit} & 51 M & 10.9 G & 83.3 \\
&TNT-B \cite{han2021transformer} & 66 M & 14.1 G & 82.8 \\
&BoTNet-S1-59 \cite{srinivas2021bottleneck} & 34 M & 7.3 G & 81.7 \\
&Focal-S \cite{Yang2021FocalAF} & 51 M & 9.4 G & 83.6 \\
&RSB-101 \cite{wightman2021resnet} & 45 M & 7.9 G & 81.3 \\
&Shuffle-B \cite{huang2021shuffle} & 88 M & 15.6 G & 84.0 \\
&Deep-ViT-L \cite{zhou2021deepvit} & 55 M & 12.5 G & 83.1 \\
&PoolFormer-M36 \cite{yu2022metaformer} & 56 M & 8.8 G & 82.1 \\
&MetaFormer \cite{yu2022metaformer} & 57 M & 12.8 G & 84.5 \\
&PVTv2-B3 \cite{wang2022pvt} & 45 M & 6.9 G & 83.2 \\
&MOA-S \cite{patel2022aggregating} & 39 M & 9.4 G & 83.5 \\
&ConvNeXt-S \cite{Liu2022ACF} & 50 M & 8.7 G & 83.1 \\
&MSG-S \cite{fang2022msg} & 56 M & 8.4 G & 83.4 \\
&MPViT-B \cite{lee2022mpvit} & 75 M & 16.4 G & 84.3 \\
&RepLKNet-31B \cite{ding2022scaling} & 79 M & 15.3 G & 83.5 \\
&VAN-B4 \cite{guo2023visual} & 60 M & 12.2 G & 84.2 \\
&Next-ViT-L \cite{li2022next} & 58 M & 10.8 G & 83.6 \\
&DilateFormer \cite{jiao2023dilateformer} & 47 M & 10.0 G & 84.4 \\
&CrossFormer-L \cite{wang2023crossformer++} & 92 M & 16.1 G & 84.0 \\
&Swin-S \cite{liu2021swin} & 50 M & 8.7 G & 83.0 \\
&\cellcolor{RowColor}{FaViT-B3 (Ours)} & \cellcolor{RowColor}48 M & \cellcolor{RowColor}8.5 G & \cellcolor{RowColor}83.4 \\
\bottomrule
\end{tabular}
\end{table}

\noindent\textbf{Data and setups.}
We evaluate the proposed FaViT variants on the classification dataset ImageNet-1K \cite{russakovsky2015imagenet}. For a fair comparison, we follow the same training strategies with priors \cite{wang2022pvt}. We use the AdamW optimizer with $300$ epochs including the initial $10$ warm-up epochs and the final $10$ cool-down epochs. A cosine decay learning rate scheduler is applied, reducing the learning rate by a factor of $10$ every $30$ epochs, beginning from a base learning rate of $0.001$. Weight decay is set to $0.05$, and input images are resized to $224 \times 224$. Data augmentations and regularization methods align with those employed in \cite{wang2022pvt}. We use a mini-batch size of 128 samples and leverage the computational power of 8 NVIDIA 3090 GPUs for training.

\noindent\textbf{Main results.}
Table \ref{classification} demonstrates the remarkable performance of the proposed FaViT variants. Specifically, the lightweight FaViT-B0 and FaViT-B1 achieve classification Top-1 accuracies of $71.5\%$ and $79.4\%$ respectively, which outperforms previous state-of-the-art models with similar numbers of parameters. Moreover, when compared to the baseline Swin Transformer \cite{liu2021swin}, FaViT variants consistently exhibit superior performance with fewer parameters and FLOPs. One striking example is FaViT-B2, which attains an impressive accuracy of $82.3\%$, surpassing Swin-T by $1\%$ while significantly reducing the parameters by $14\%$. This exceptional accuracy and efficiency can be attributed to FaViT's unique ability to model long-range dependency at the local window cost.

\begin{figure}[!t]
\centering
\includegraphics[width=3.5in]{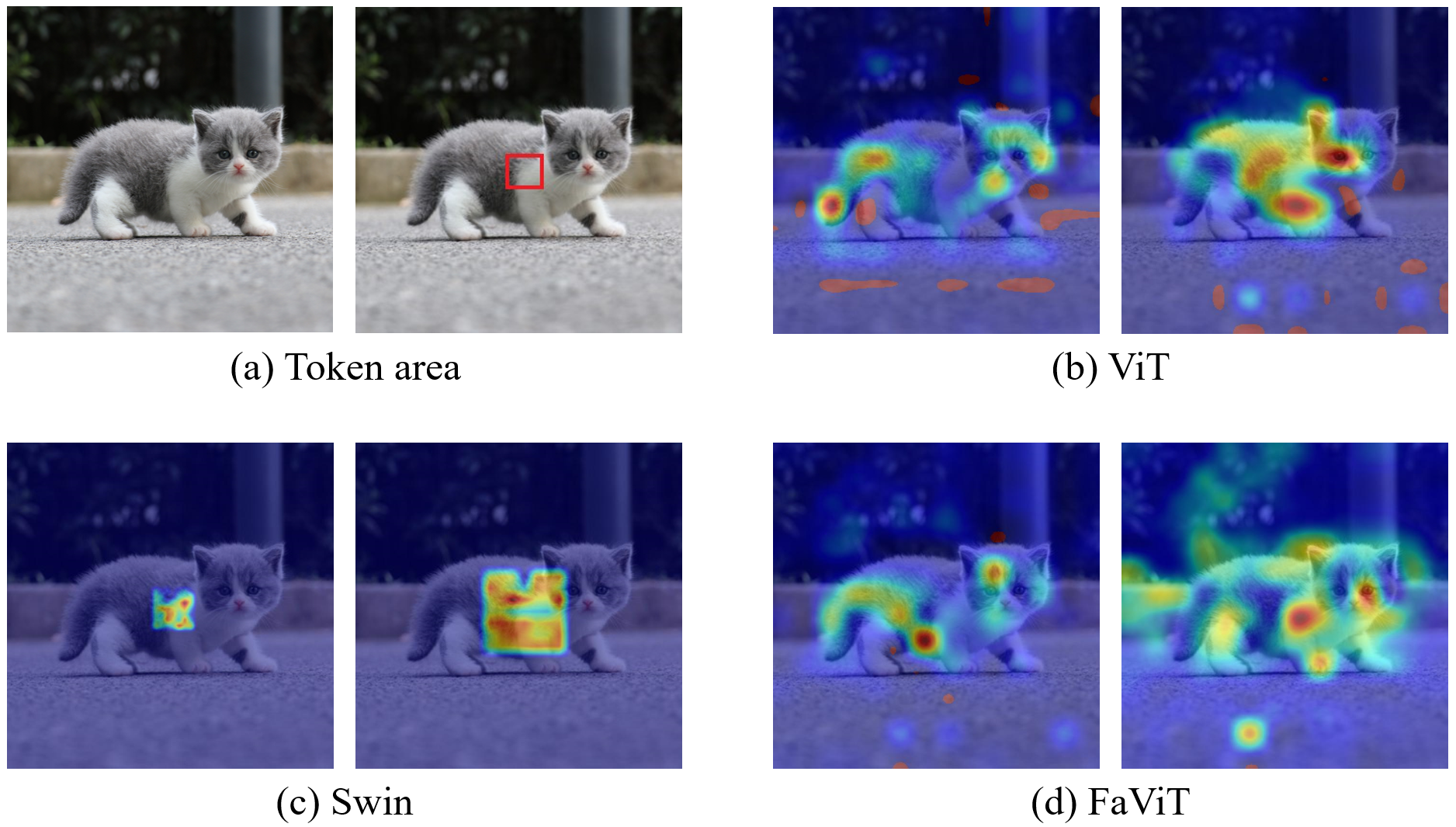}
\caption{Visualizing the attention spans for a given token area (red box) by different models. (a) Original image with selected area that is marked by a red box. (b) ViT has global attention span but high computation cost. (c) Swin restricts the attention span to a local window. (d) FaViT achieves global attention span at local window cost.}
\label{fig:field}
\end{figure}

\noindent\textbf{Visualization.}
FaViT achieves high classification accuracy due to its attention span across both short and long ranges. As shown in Figure \ref{fig:field}, we visualize the attention spans from the first and the second stages of different models for a given token area. The baseline model Swin transformer \cite{liu2021swin} focuses only on the small neighborhood of the token, which leads to degraded accuracy and robustness. In contrast, ViT \cite{dosovitskiy2020image} has a longer-range attention span owing to the adopted global attention mechanism, but at the cost of quadratically increased computational complexity with the increase of input image spatial resolution. The proposed FaViT well takes the complementary strengths of both models.
It enjoys a large attention span similar to ViT while maintaining a computational cost equivalent to the local window-based Swin Transformer. This balanced approach achieves an ideal trade-off between the capability to model long-range dependency and computational cost.

\begin{figure}[!t]
\centering
\includegraphics[width=3.5in]{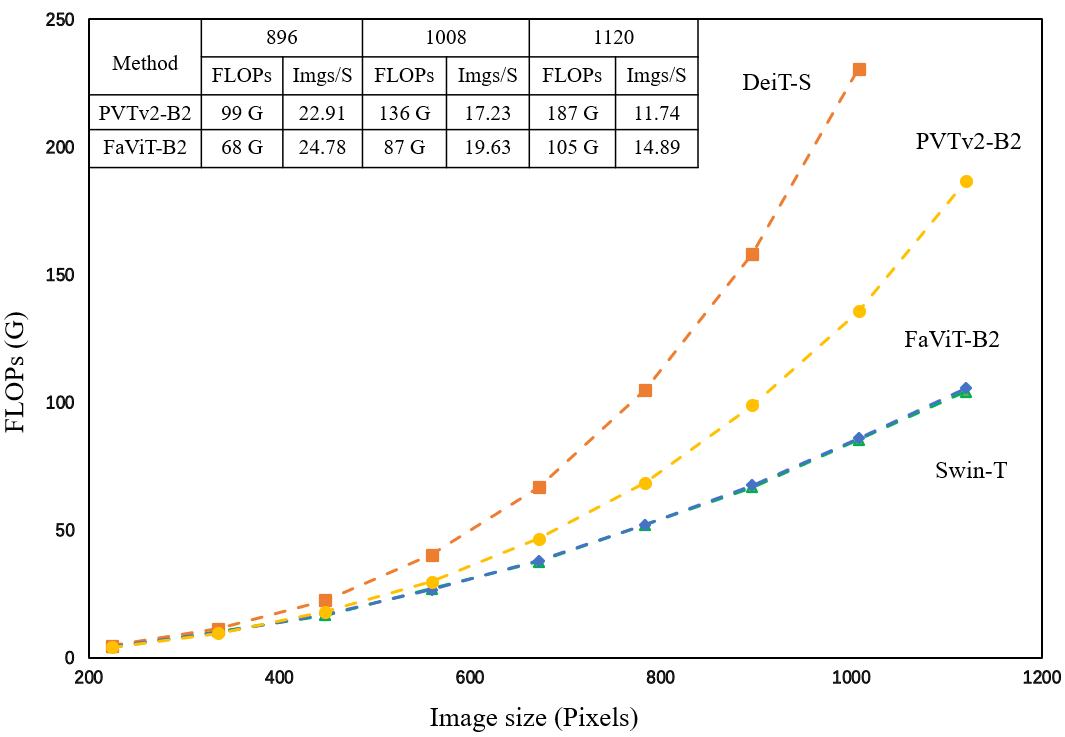}
\caption{Diagram for the input image size vs FLOPs. The FLOPs and Throughputs of PVTv2 and FaViT at high resolution are listed at the upper left corner.}
\label{fig:flopvs}
\end{figure}

\noindent\textbf{Efficiency analysis.}
We compare the computational complexity of the proposed FaViT with other priors in Figure \ref{fig:flopvs} by reporting the number of FLOPs under varying input image sizes. When the input image size is small, \textit{e.g.}, $224 \times 224$, all the models have a comparable number of FLOPs. However, as the input image size increases, we observe significant differences. The conventional DeiT \cite{touvron2021training} suffers a dramatic rise in FLOPs as the image size grows, as its computational complexity is quadratic with respect to the image size. This limitation restricts its applicability in real-world scenarios. Although PVTV2 \cite{wang2022pvt} has managed to slow down the rate of FLOP increase, it still fundamentally exhibits quadratic growth in computational complexity. In contrast, FaViT's FLOPs increase linearly with the image size due to its adoption of a local window-based self-attention strategy. This linear growth makes FaViT a more scalable and efficient choice for handling larger input images.

The computational superiority of FaViT becomes particularly evident when dealing with large input images. Specifically, when the image size reaches $1000 \times 1000$ pixels, FaViT-B2 exhibits a substantial advantage in terms of FLOPs compared to DeiT-S and PVTv2-B2, with only $25\%$ and $50\%$ of their computational cost, respectively. In comparison with Swin-T \cite{liu2021swin}, FaViT-B2 achieves the same computational cost while delivering higher accuracy. Furthermore, we report the throughputs (imgs/s) at high image sizes for PVTv2-B2 and FaViT-B2. The proposed FaViT is more efficient and the advantage becomes more pronounced as the size increases.

\subsection{Object Detection}

\noindent\textbf{Data and setups.}
We evaluate the proposed FaViT variants in the challenging tasks of object detection and instance segmentation on the COCO 2017 dataset \cite{lin2014microsoft}. To ensure a comprehensive evaluation, we employ two well-established frameworks, RetinaNet \cite{lin2017focal} and Mask R-CNN \cite{he2017mask}, and integrate FaViT variants as backbones. We initialize the models with weights pre-trained on the ImageNet-1K dataset and fine-tune them using the AdamW optimizer. We set the initial learning rate at $0.0001$ and apply a weight decay of $0.05$. Our training is conducted with a batch size of $16$, adhering to a consistent $1\times$ training schedule, which amounts to 12 epochs. Additionally, we adopt an image resizing strategy as priors \cite{liu2021swin}, where the shorter side is resized to 800 pixels while ensuring that the longer side remains under 1333 pixels. The rest of the hyperparameters are consistent with the Swin transformer \cite{liu2021swin} to ensure a fair comparison.

\begin{table*}[!t]
\caption{Object detection and instance segmentation results on COCO 2017. All models are pre-trained on ImageNet-1K and fine-tuned with $1\times $ schedule. The number of parameters at an input resolution of $1280 \times 800$ is reported (\#Param).\label{tab:detection}}
\centering
\renewcommand\arraystretch{1.2}
\begin{tabular}{l|ccccccc|ccccccc}
\toprule
\multicolumn{1}{c|}{\multirow{2}{*}{Method}}  & \multicolumn{7}{c|}{RetinaNet $1\times $} & \multicolumn{7}{c}{Mask R-CNN $1\times $}\\
\cmidrule(lr){2-8}\cmidrule(lr){9-15}
\multicolumn{1}{c|}{} &{\#Param} &$\rm{AP}$ &$\rm{AP}_{50}$ &$\rm{AP}_{75}$ &$\rm{AP}_{S}$ &$\rm{AP}_{M}$ &$\rm{AP}_{L}$ &{\#Param} &$\rm{AP}^{b}$ &$\rm{AP}^{b}_{50}$ &$\rm{AP}^{b}_{75}$ &$\rm{AP}^{m}$ &$\rm{AP}^{m}_{50}$ &$\rm{AP}^{m}_{75}$\\
\midrule
PVTv2-B0 \cite{wang2022pvt} &13 M &37.2 &57.2 &39.5 &\textbf{23.1} &40.4 &\textbf{49.7} &24 M &\textbf{38.2} &\textbf{60.5} &40.7 &\textbf{36.2} &\textbf{57.8} &\textbf{38.6}\\
\rowcolor{RowColor}FaViT-B0 (Ours)       &\textbf{12 M} &\textbf{37.4} &\textbf{57.2} &\textbf{39.8} &22.9 &\textbf{40.6} &49.5 &\textbf{23 M} &37.9 &59.6 &\textbf{41.0} &35.4 &56.6 &37.8\\
\midrule
ResNet18 \cite{he2016deep} &\textbf{21 M} &31.8 &49.6 &33.6 &16.3 &34.3 &43.2 &\textbf{31 M} &34.0 &54.0 &36.7 &31.2 &51.0 &32.7\\
PVTv2-B1 \cite{wang2022pvt} &24 M &41.2 &\textbf{61.9} &43.9 &25.4 &44.5 &54.3 &34 M &41.8 &64.3 &45.9 &38.8 &61.2 &41.6\\
\rowcolor{RowColor}  FaViT-B1 (Ours) &23 M &\textbf{41.4} &61.8 &\textbf{44.0} &\textbf{25.7} &\textbf{44.9} &\textbf{55.0} &33 M &\textbf{42.4} &\textbf{64.4} &\textbf{46.3} &\textbf{38.9} &\textbf{61.2} &\textbf{41.8}\\
\midrule
Twins-S \cite{chu2021twins}  &\textbf{34 M} &43.0 &64.2 &46.3 &28.0 &46.4 &57.5 &\textbf{44 M} &43.4 &66.0 &47.3 &40.3 &63.2 &43.4\\
Swin-T \cite{liu2021swin} &39 M &41.5 &62.1 &44.2 &25.1 &44.9 &55.5 &48 M &42.2 &64.6 &46.2 &39.1 &61.6 &42.0\\
\rowcolor{RowColor} FaViT-B2 (Ours) &35 M &\textbf{44.4} &\textbf{65.0} &\textbf{47.7} &\textbf{27.7} &\textbf{48.2} &\textbf{58.8} &45 M &\textbf{45.4} &\textbf{67.1} &\textbf{49.4} &\textbf{41.0} &\textbf{64.0} &\textbf{44.1}\\
\midrule
PVTv1-M \cite{wang2021pyramid}  &\textbf{54 M} &41.9 &63.1 &44.3 &25.0 &44.9 &57.6 &\textbf{64 M} &42.0 &64.4 &45.6 &39.0 &61.6 &42.1\\
Swin-S \cite{liu2021swin} &60 M &44.5 &65.7 &47.5 &27.4 &48.0 &59.9 &69 M &44.8 &66.6 &48.9 &40.9 &63.4 &44.2\\
\rowcolor{RowColor} FaViT-B3 (Ours) &59 M &\textbf{46.0} &\textbf{66.7} &\textbf{49.1} &\textbf{28.4} &\textbf{50.3} &\textbf{62.0} &68 M &\textbf{47.1} &\textbf{68.0} &\textbf{51.4} &\textbf{42.7} &\textbf{65.9} &\textbf{46.1}\\
\bottomrule
\end{tabular}
\end{table*}

\noindent\textbf{Main results.}
In Table \ref{tab:detection}, we present the performance results of FaViT-B0 to B3 across various size levels. Impressively, all variants consistently deliver state-of-the-art results. In comparison to the baseline Swin transformer \cite{liu2021swin}, FaViT showcases superior performance in both object detection and instance segmentation tasks. Specifically, FaViT-B2 showcases substantial improvements, with metrics such as AP and $AP_L$ witnessing gains of $2.9\%$ and $3.3\%$, respectively. For instance segmentation, FaViT-B2 also achieves higher accuracy and significantly improves $AP^m$ by $1.9\%$.

\begin{figure}[!t]
\centering
\includegraphics[width=3.5in]{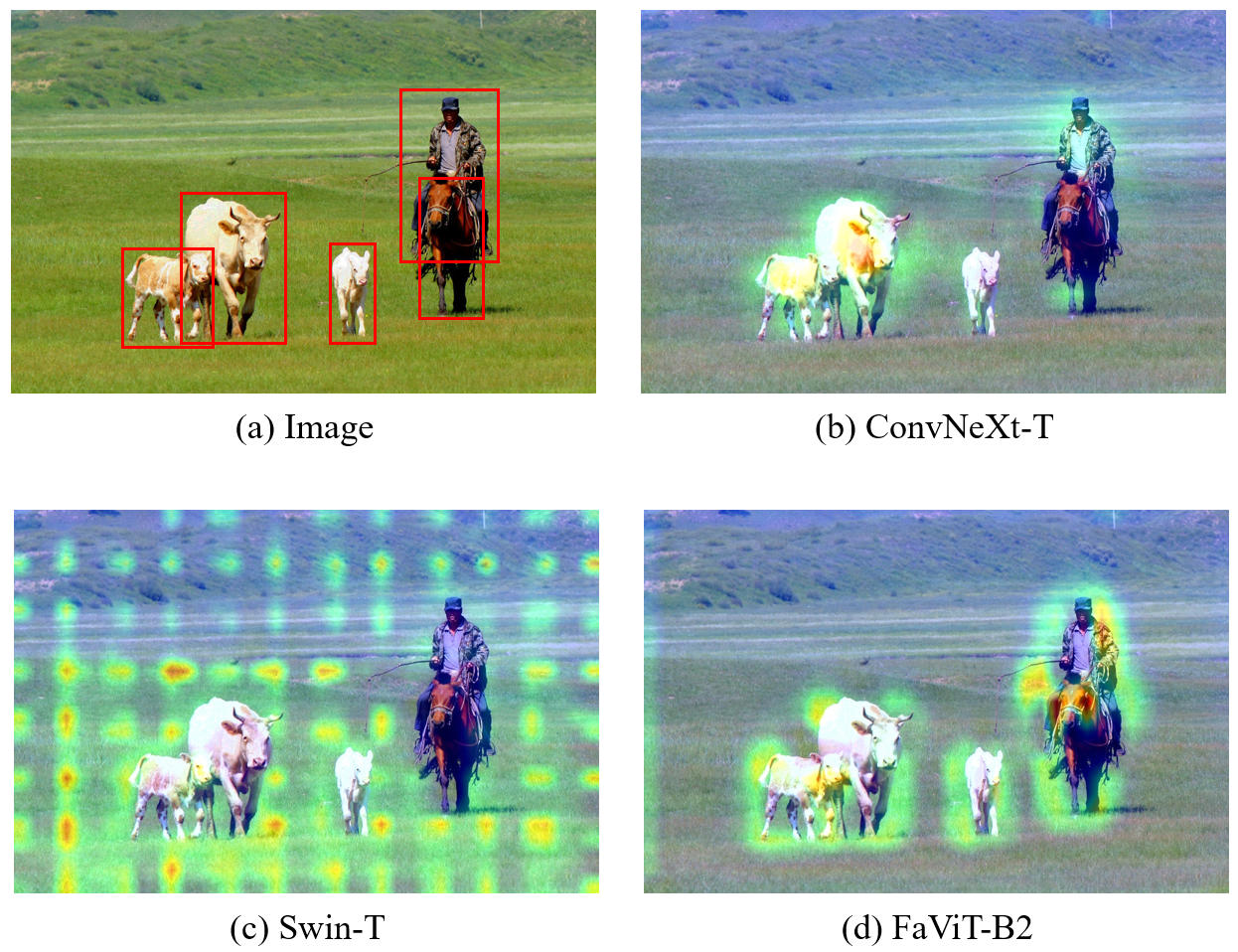}
\caption{Visualizing attention maps for detecting multiple objects. (a) Multiple targets to be detected in the original image are marked with red boxes. (b) ConvNeXt-T loses small target when detecting multiple targets. (c) Swin-T is severely affected by local windows. (d) FaViT-B2 has excellent multi-target detection capability. }
\label{fig:mul}
\end{figure}

\noindent\textbf{Visualization.}
Due to the long-range dependency, FaViT enjoys advantages in detecting various scale objects. We visualize the attention heatmap of FaViT-B2 under the multi-object challenge in Figure \ref{fig:mul}. To offer a comparative analysis, we juxtapose FaViT-B2 with ConvNeXt-T \cite{Liu2022ACF} and Swin-T, two models of similar size. ConvNeXt-T appears less adept at simultaneously focusing on multiple objects, potentially causing it to overlook individual targets. Swin transformer's attention heatmap exhibits a grid-like structure, which is relatively messy. In contrast, FaViT's attention distribution is more uniform across the entire object space, allowing it to simultaneously capture the positions and contours of multiple objects with precision. This heatmap comparison serves as compelling evidence of FaViT's superiority.

\subsection{Semantic Segmentation}
\noindent\textbf{Data and setups.}
To evaluate the performance of the proposed FaViT variants in semantic segmentation, we employ the ADE20K dataset \cite{zhou2019semantic} and follow a benchmark framework, Semantic FPN \cite{kirillov2019panoptic}, to ensure a fair comparison. Our training strategy aligns with established practices \cite{liu2021swin}, encompassing 160K iterations, utilization of the AdamW optimizer, an initial learning rate of $0.0002$, a weight decay of $0.0001$, and a batch size of $16$.

\noindent\textbf{Main results.}
Table \ref{tab:segmentation} depicts that our FaViT variants of different model sizes consistently outperform their corresponding Swin transformer \cite{liu2021swin} counterparts when using Semantic FPN for semantic segmentation. For example, with fewer parameters and FLOPs, FaViT-B2/B3 are at least $2\%$ higher than Swin-T/S. Noticeably, FaViT shows superior segmentation accuracy across all model sizes.

\begin{table}[!t]
\caption{Segmentation results on ADE20K.\label{tab:segmentation}}
\centering
\renewcommand\arraystretch{1.2}
\setlength{\tabcolsep}{10pt}
\begin{tabular}{l|ccc}
\toprule
\multicolumn{1}{c|}{Method} & \#Param & FLOPs & mIoU (\%) \\
\midrule
PVTv2-B0 \cite{wang2022pvt} & 8 M & 25.0 G & 37.2 \\ 
\rowcolor{RowColor}{FaViT-B0 (Ours)} & \textbf{7 M} & \textbf{24.6 G} & \textbf{37.2} \\ 
\midrule
ResNet18 \cite{he2016deep} & 16 M & 32.2 G & 32.9 \\
PVTv1-T \cite{wang2021pyramid} & \textbf{13 M} & \textbf{31.9 G} & 35.1 \\
EFormer-L1 \cite{Li2022EfficientFormerVT} & 16 M & 33.0 G & 38.9 \\
\rowcolor{RowColor} FaViT-B1 (Ours) & 17 M & 33.9 G & \textbf{42.0} \\
\midrule
Twins-S \cite{chu2021twins} & \textbf{28 M} & \textbf{37.5 G} & 43.2 \\
Swin-T \cite{liu2021swin} & 32 M & 46.0 G & 41.5 \\
\rowcolor{RowColor} FaViT-B2 (Ours) & 29 M & 45.2 G & \textbf{45.0} \\
\midrule
CAT-B \cite{Lin2022CATCA} & 55 M & 76.9 G & 44.9 \\
Swin-S \cite{liu2021swin} & 53 M & 70.0 G & 45.2 \\
\rowcolor{RowColor} FaViT-B3 (Ours) & \textbf{52 M} & \textbf{66.7 G} & \textbf{47.2} \\
\bottomrule
\end{tabular}
\end{table}

\subsection{Robustness Analysis}
We evaluate the robustness of FaViT variants to image corruptions, label noise, and class imbalance.

\begin{table*}[!t]
\caption{Robustness against image corruptions on ImageNet-C.\label{robust}}
\centering
\renewcommand\arraystretch{1.2}
\setlength{\tabcolsep}{2pt}
\begin{tabular}{l|c|c|cccc|cccc|cccc|cccc}
\toprule
\multicolumn{1}{c|}{\multirow{2}{*}{Method}}  & \multirow{2}{*}{\#Param} & \multirow{2}{*}{Retention} & \multicolumn{4}{c|}{Blur (Acc \%)} & \multicolumn{4}{c|}{Noise (Acc \%)} & \multicolumn{4}{c|}{Digital (Acc \%)} &\multicolumn{4}{c}{Weather (Acc \%)}\\
\cmidrule(lr){4-7}\cmidrule(lr){8-11}\cmidrule(lr){12-15}\cmidrule(lr){16-19}
\multicolumn{1}{c|}{} & & & Motion & Fafoc & Glass & Gauss & Gauss & Impul & Shot & Speck & Contr & Satur & JEPG & Pixel & Bright & Snow & Fog & Frost\\
\midrule
\multicolumn{19}{c}{Mobile Setting (\textless 15M)}\\
\midrule
MobileNetV2 \cite{2018MobileNetV2}  &4 M &49.2 & 33.4 & 29.6 & 21.3 &32.9 & 24.4 & 21.5 & 23.7 &32.9 & 57.6 & 49.6 & 38.0 &62.5 & 28.4 & 45.2 & 37.6 & 28.3\\
EfficientNet-B0 \cite{tan2019efficientnet}  &5 M &54.7 & 36.4 & 26.8 & \textbf{26.9} &\textbf{39.3} & \textbf{39.8} & \textbf{38.1} & \textbf{47.1} &39.9 & \textbf{65.2} & 58.2 & \textbf{52.1} &\textbf{69.0} & 37.3 & \textbf{55.1} & 44.6 & 37.4\\
PVTv2-B0 \cite{wang2022pvt} &3 M &58.9 & 30.8 & 24.9 & 24.0 &35.8 & 33.1 & 35.2 & 44.2 &50.6 & 59.3 & 50.8 & 36.6 &61.9 & 38.6 & 50.7 & 45.9 & 41.8\\
ResNet18 \cite{he2016deep} &\textbf{11 M} &32.8 &29.6 &28.0 &22.9 &32.0 &22.7 &17.6 &20.8 &27.7 &30.8 &52.7 &46.3 &42.3 &58.8 &24.1 &41.7 &28.2\\
PVTv2-B1 \cite{wang2022pvt} &13 M &65.4 & 45.7 & 41.3 & 30.5 &43.9 & 48.1 & 46.2 & 46.6 &55.0 & 57.6 & 68.6 & 59.9 &50.2 & 71.0 & 49.8 & 56.8 & 53.0\\
\midrule
\rowcolor{RowColor} FaViT-B0 (Ours) & \textbf{3 M} & \textbf{59.2}& \textbf{38.1}& \textbf{31.6}& 24.8& 37.4& 38.3& 35.6& 39.9& \textbf{45.2}& 47.9& \textbf{60.8}& 51.6& 38.9& \textbf{63.2}& 38.5& \textbf{44.6}& \textbf{42.5}\\
\rowcolor{RowColor} FaViT-B1 (Ours) & 13 M & \textbf{68.1} & \textbf{48.2} & \textbf{43.2} & \textbf{30.7} & \textbf{45.6} & \textbf{53.8} & \textbf{52.4} & \textbf{52.6} & \textbf{58.7} & \textbf{59.6} & \textbf{70.1} & \textbf{61.7} & \textbf{53.5} & \textbf{72.1} & \textbf{50.9} & \textbf{57.1} & \textbf{54.7} \\
\midrule
\multicolumn{19}{c}{GPU Setting (20M+)}\\
\midrule
ResNet50 \cite{he2016deep}  &25 M &62.5 & 42.1 & 40.1 & 27.2 &42.2 & 42.2 & 36.8 & 41.0 &50.3 & 51.7 & 69.2 & 59.3 &51.2 & 71.6 & 38.5 & 53.9 & 42.3\\
ViT-S \cite{dosovitskiy2020image} &25 M &67.6 & 49.7 & 45.2 & \textbf{38.4} &48.0 & 50.2 & 47.6 & 49.0 &57.5 & 58.4 & 70.1 & 61.6 &57.3 & 72.5 & 51.2 & 50.6 & 57.0\\
DeiT-S \cite{touvron2021training} &\textbf{22 M} &72.6 & 52.6 & 48.9 & 38.1 & 51.7 & 57.2 & 55.0 & 54.7 & 60.8 & 63.7 & 71.8 & 64.0 & 58.3 & 73.6 & 55.1 & 61.1 & 60.7\\
PVTv1-S \cite{wang2021pyramid} &25 M &66.9 & 54.3 & 48.4 & 34.7 &46.4 & 51.7 & 51.7 & 50.0 &55.8 & 57.6 & 69.4 & 60.7 &53.7 & 49.5 & \textbf{62.3} & 55.2 & 53.1\\
PVTv2-B2 \cite{wang2022pvt} &25 M &71.5 & 54.3 & 48.4 & 34.7 &50.7 & 61.2 & 60.7 & 59.5 &\textbf{64.5} & \textbf{65.5} & \textbf{73.5} & 65.5 &58.8 & \textbf{75.2} & 56.7 & \textbf{67.8} & \textbf{62.7}\\
Swin-T \cite{liu2021swin} &29 M &66.8 &49.5 & 45.0 & 31.7 &47.6 &54.7 & 51.6 &52.6 &58.4 &62.1 & 71.4 & 62.2 &54.4 & 73.4 & 60.0 & 64.7 & 60.2\\
\midrule
\rowcolor{RowColor} FaViT-B2 (Ours) & 25 M & \textbf{73.4} & \textbf{55.9} & \textbf{49.8} & 34.8 & \textbf{51.7} & \textbf{62.6} & \textbf{62.1} & \textbf{61.2} & 65.3 & 64.9 & 73.3 & \textbf{66.1} & \textbf{64.8} & 75.0 & 55.3& 62.9 & 59.5 \\
\bottomrule
\end{tabular}
\end{table*}

\noindent\textbf{Robustness against image corruptions.}
We evaluate the robustness of our proposed FaViT variants on the ImageNetC dataset \cite{Hendrycks2019Benchmarking}, which includes various corrupted images with effects like blur, natural noise, digital noise, and severe weather conditions. For a fair comparison, all models are pre-trained on ImageNet-1K without further fine-tuning \cite{zhou2022understanding}. The results in Table \ref{robust} demonstrate that FaViT exhibits remarkable robustness, both under Mobile and GPU settings, surpassing CNN and transformer-based priors. This enhanced robustness likely stems from its superior representation capabilities. To better gauge model robustness, we introduce a new metric called accuracy retention (Retention), designed to mitigate the influence of a model's capability to represent on clean images dataset. This metric quantifies the ratio between a model's accuracy on corrupted images and its accuracy on clean images, providing insight into how well a model preserves accuracy when tested on corrupted data. 

For instance, consider FaViT-B1, which achieves an accuracy of $54.1\%$ on ImageNet-C and $79.4\%$ on ImageNet. Its accuracy retention is $68.1\%$, indicating that it preserves $68.1\%$ of its accuracy when tested on corrupted images. In comparison, PVTv2-B1 \cite{wang2022pvt}, while achieving a similar accuracy on clean images, has an accuracy retention of $65.4\%$. This suggests that FaViT consistently outperforms PVTv2 when it comes to preserving accuracy on corrupted images. Notably, FaViT-B2 significantly outperforms Swin-T \cite{liu2021swin} by $6.6\%$ in accuracy retention and demonstrates superior performance across various types of image corruptions. These results underscore FaViT's success in capturing long-range contextual information, a crucial factor in enhancing robustness against image corruptions.

\begin{table}[!t]
\caption{Robustness against label noise on Clothing1M and Webvision.\label{noise}}
\centering
\renewcommand\arraystretch{1.2}
\setlength{\tabcolsep}{3pt}
\begin{tabular}{l|c|c|cc}
\toprule
\multicolumn{1}{c|}{\multirow{2}{*}{Method}}  & \multirow{2}{*}{\#Param} & \multicolumn{1}{c|}{Clothing1M} &  \multicolumn{2}{c}{Webvision}\\
\cmidrule(lr){3-3}\cmidrule(lr){4-5}
\multicolumn{1}{c|}{} & &Test Acc (\%) &Top-1 Acc (\%) &Top-5 Acc (\%) \\
\midrule
PVTv1-S \cite{wang2021pyramid} &25 M &68.83 &60.08 &81.84\\
PVTv2-B2 \cite{wang2022pvt} &25 M &69.89 &65.28 &85.72\\
Shunted-S \cite{ren2021shunted} &\textbf{22 M} &70.04 &67.44 &\textbf{86.24}\\
Swin-T \cite{liu2021swin} &29 M &69.12 &60.84 &82.48\\
\midrule
\rowcolor{RowColor} FaViT-B2 (Ours) &25 M &\textbf{70.82} &\textbf{67.72} &85.80\\
\bottomrule
\end{tabular}
\end{table}

\noindent\textbf{Robustness against label noise.}
We assess the robustness of FaViT against real-world label noise using the Clothing1M \cite{Tong2015Learning} and WebVision \cite{Li2017WebVisionDV} datasets. To ensure a fair comparison, we adopt a consistent training strategy as \cite{Li2020DivideMixLW} across all models. Table \ref{noise} presents the results, with FaViT-B2 consistently achieving the highest accuracy on both datasets. Remarkably, FaViT-B2 attains a top-1 accuracy of $67.72\%$ on WebVision, outperforming the baseline Swin-T \cite{liu2021swin} by a substantial margin of $6.9\%$. These findings underscore the significant robustness of FaViT in the presence of real-world label noise.

\begin{table}[!t]
\caption{Robustness against class imbalance on iNaturalist 2018.\label{class}}
\centering
\renewcommand\arraystretch{1.2}
\setlength{\tabcolsep}{15pt}
\begin{tabular}{l|cc}
\toprule
\multicolumn{1}{c|}{Method}       & FLOPs & Top-1 Acc (\%) \\
\midrule
ResMLP-12 \cite{Touvron2022ResMLPFN}  &3.0 G        & 60.2 \\
Inception-V3 \cite{Horn2018TheIS}  &2.5 G  & 60.2 \\
LeViT-192 \cite{Graham2021LeViTAV}  &\textbf{0.7 G}      & 60.4 \\
ResNet-50 \cite{Cui2019ClassBalancedLB}     &4.1 G     & 64.1 \\
\midrule
\rowcolor{RowColor} FaViT-B1 (Ours)  &2.4 G & \textbf{64.2} \\
\bottomrule
\end{tabular}
\end{table}

\noindent\textbf{Robustness against class imbalance.}
We test the model robustness against class imbalance on the long-tailed iNaturalist dataset \cite{Horn2018TheIS}. All models are pre-trained on ImageNet-1K and fine-tuned for $100$ epochs with an initial learning rate of $0.0001$. Table \ref{class} presents the results, with FaViT-B1 achieving the highest accuracy compared to other models. This underscores the excellent robustness of FaViT when dealing with long-tailed data. The outcomes from these experiments demonstrate the strong robustness of FaViT against data corruptions and biases, showcasing its potential for real-world applications. Table \ref{class} also illustrates that FaViT performs competitively in terms of FLOPs compared to the state-of-the-art. Table \ref{class} also illustrates that FaViT performs competitively in terms of FLOPs compared to the state-of-the-art.

\subsection{Ablation Study}
We perform ablation studies on CIFAR100 \cite{krizhevsky2009learning}. All model variants are trained from scratch for $100$ epochs with an initial learning rate of $0.001$. The rest of the training strategy is consistent with Section \ref{IC}.

\noindent\textbf{Effectiveness of FaSA.}
We evaluate the effectiveness and generalizability of the proposed factorization self-attention (FaSA) mechanism by integrating it to other transformer-based backbones. Concretely, we simply replace the original self-attention in the Swin transformer \cite{liu2021swin} and PVTv2 \cite{wang2022pvt} with FaSA while keeping the rest of the network architecture unchanged. Table \ref{tab:ablation1} demonstrates that FaSA consistently improves the performance of various backbones while simultaneously reducing the number of parameters and FLOPs. In particular, it significantly enhances the accuracy of Swin-T and PVTv2-B1 by $4.3\%$ and $2.3\%$, respectively, while reducing the number of parameters of PVTv2-B1 by $18\%$. The results provide clear evidence of the superiority of FaSA compared to other popular self-attention mechanisms. 

\begin{table}[!t]
\caption{Apply FaSA to other frameworks.\label{tab:ablation1}}
\centering
\renewcommand\arraystretch{1.2}
\setlength{\tabcolsep}{10pt}
\begin{tabular}{l|ccc}
\toprule
\multicolumn{1}{c|}{Method} & \#Param & FLOPs & Acc (\%)\\
\midrule
Swin-T \cite{liu2021swin} & 28 M     & 4.4 G      & 61.8 \\
\rowcolor{RowColor} Swin-FaSA-T     & \textbf{26 M}     & \textbf{4.0 G}      & \textbf{66.1}  \\ 
\midrule
PVTv2-B0 \cite{wang2022pvt} & 4 M     & 0.6 G      & 63.5 \\
\rowcolor{RowColor} PVTv2-FaSA-B0   & \textbf{3 M}      & \textbf{0.6 G}      & \textbf{64.0} \\
\midrule
PVTv2-B1 \cite{wang2022pvt} & 14 M     & 2.1 G      & 70.8 \\
\rowcolor{RowColor} PVTv2-FaSA-B1    & \textbf{11 M}      & \textbf{2.1 G}      & \textbf{73.1}  \\
\bottomrule
\end{tabular}
\end{table}

\begin{table}[!t]
\caption{Reduce mixed to single fine-grained.\label{tab:ablation2}}
\centering
\renewcommand\arraystretch{1.2}
\setlength{\tabcolsep}{12.5pt}
\begin{tabular}{l|ccc}
\toprule
\multicolumn{1}{c|}{Method} & \#Param & FLOPs & Acc (\%)\\
\midrule
FaSA-low           & 3 M     & 0.6 G   &64.9    \\
FaSA-high      & 3 M     & 0.6 G    &64.5   \\
\midrule
\rowcolor{RowColor} FaSA    & \textbf{3 M}      & \textbf{0.6 G}    &\textbf{65.2}   \\
\bottomrule
\end{tabular}
\end{table}

\noindent\textbf{Impact of dilation rate set.}
In FaSA, we focus on aggregating mixed-grained information captured by grouped features. To further investigate the impact of this mixed-grained aggregation, we design two additional models, FaSA-low and FaSA-high, which each utilize only a single level of granularity information based on FaViT-B0. FaSA-low represents a configuration where the dilation rate for each group is set to 1, resulting in the extracted queries having the lowest granularity information. FaSA-high, on the other hand, is configured to have local windows similar in size to the feature map, emphasizing higher granularity information. Table \ref{tab:ablation2} presents the results, showing that the proposed FaSA consistently outperforms FaSA-low and FaSA-high, indicating the effectiveness of aggregating mixed-grained information in FaSA.

\begin{table}[!t]
\caption{Imapct of global features.\label{tab:ablation3}}
\centering
\renewcommand\arraystretch{1.2}
\setlength{\tabcolsep}{6pt}
\begin{tabular}{l|c|cccc}
\toprule
\multicolumn{1}{c|}{\multirow{2}{*}{Method}}    &\multirow{2}{*}{Acc (\%)}       & \multicolumn{4}{c}{Param w.r.t image size} \\
\cmidrule(lr){3-6}
&     & 448  & 672    & 896    & 1120 \\
\midrule
FaViT-B$\rm{2}_{0}$     & 75.7    & 18 G    & \textbf{41 G}  &\textbf{72 G}  & \textbf{113 G}   \\
FaViT-B$\rm{2}_{1/4}$   & 75.7    & 18 G    & 43 G  &80 G  & 135 G \\
\rowcolor{RowColor} FaViT-B$\rm{2}_{1/8}$   & \textbf{75.8}    & \textbf{18 G}    & 42 G  &76 G  & 124 G  \\
FaViT-B$\rm{2}_{1/16}$  & 75.8    & 18 G    & 41 G  &74 G  & 118 G   \\
\bottomrule
\end{tabular}
\end{table}

\noindent\textbf{Optimization structure with global features.}
The proposed FaSA introduces dilation rates to increase the local window size and model long-range but not global dependency. However, we argue that introducing an appropriate amount of global features may help to improve model performance. To investigate this, we split part of the channels and extract global features from it using the method in \cite{wang2022pvt}. Table \ref{tab:ablation3} demonstrates that introducing global features improves model performance while increasing computational cost. However, this improvement comes at the cost of increased computational complexity. To strike a balance between performance and cost, we extract global features from 1/8 of the channels, while FaSA handles the rest. This configuration achieves an ideal trade-off between improved performance and manageable computational cost.

\begin{table}[!t]
\caption{Comparison of different operators used for cross-window fusion on CIFAR100.\label{tab:operators}}
\centering
\renewcommand\arraystretch{1.2}
\setlength{\tabcolsep}{6pt}
\begin{tabular}{l|ccc}
\toprule
\multicolumn{1}{c|}{Operator} & \#Param & FLOPs & Top-1 Acc (\%) \\
\midrule
Pointwise Convolution & 3.5 M & 0.6 G & 68.6 \\
Linear Layer & 3.5 M & 0.6 G & 67.3 \\
Average Pooling & 3.4 M & 0.6 G & 68.6 \\
\rowcolor{RowColor} Maximum Pooling (FaViT-B0) & \textbf{3.4 M} & \textbf{0.6 G} & \textbf{68.9} \\
\bottomrule
\end{tabular}
\end{table}

\noindent\textbf{Operator of cross-window fusion.}
In the cross-window fusion step of FaSA, we use a symmetric aggregation function $\sigma(\cdot)$ to fuse the features of the sampled points at the same position in different windows. In practice, the fusion operator is the average pooling. In order to find the best operator, we use FaViT-B0 as a benchmark to test the performance of various operators, including pointwise convolution, linear layer, maximum pooling layer, and average pooling layer. We evaluate the classification accuracy on CIFAR100, train $100$ epochs from scratch, and other training strategies are consistent with Section \ref{IC}. Table \ref{tab:operators} shows the comparison results of various operators, where the use of learnable operators such as pointwise convolution and linear layer increases parameters, but there is no performance advantage. Furthermore, we compare two common pooling layers and finally choose the maximum pooling with better performance.

\section{Future Work}
The FaViT proposed in this paper indeed finds the trade-off between cost and performance, but there are still areas for improvement. First, compared with some existing transformers \cite{wang2022pvt,ren2021shunted}, the proposed FaViT only innovates the core attention mechanism, without a specially designed patch embedding layer and feedforward layer. These layers may not be an optimal match with FaSA, affecting performance. Second, some existing transformers \cite{zhang2022vsa, tolstikhin2021mlp, guo2022cmt} have begun to explore the use of convolution, pooling, or multi-layer perceptrons to replace self-attention. Compared to these methods, FaViT does not involve related work, but they are instructive and attractive. Therefore, our future work will focus on model optimization, designing model structures with higher efficiency and better performance. We will also further explore the feasibility of other approaches to implement attention matrix decomposition. Furthermore, we will evaluate the performance of existing methods and the proposed FaViT on high-resolution image processing, making the transformers more valuable for practical applications.

\section{Conclusion}
This paper proposes a novel factorization self-attention mechanism (FaSA) to explore the optimal trade-off between computational cost and the capability to model long-range dependency. 
We introduce a factorization operation to obtain the long-range and mixed-grained information simultaneously.
With the aid of FaSA, long-range dependency will be modeled at the local window equivalent computational cost.
Extensive experiments show that the proposed model achieves state-of-the-art performance and superior robustness. 
We hope this work will provide reference and inspiration for future research on visual transformers.

\section*{Acknowledgments}
This work was financially supported by the National Natural Science Foundation of China (No. 62101032), the Young Elite Scientist Sponsorship Program of China Association for Science and Technology (No. YESS20220448), and the Young Elite Scientist Sponsorship Program of Beijing Association for Science and Technology (No. BYESS2022167).



\bibliography{refer}

\end{document}